# Domain Adaptation of Large Language Models for Polymer-Composite Additive Manufacturing Using Retrieval-Augmented Generation and Fine-Tuning

**Saiful Islam Sagor[1], Tania Haghighi[2], Minhaj Nur Alam[2]*, Erina Baynojir Joyee[1]***

**[1]Department of Mechanical Engineering and Engineering Science, University of North Carolina at Charlotte, NC 28223, USA.**

**[2]Department of Electrical and Computer Engineering, University of North Carolina at Charlotte, NC 28223, USA.**

***Corresponding author(s). E-mail(s): ejoyee@charlotte.edu**

**Abstract**

General-purpose large language models (LLMs) often struggle to generate reliable responses in specialized engineering domains due to limited domain grounding and insufficient exposure to structured technical knowledge. This study investigates practical strategies for adapting a foundation LLM to the additive manufacturing (AM) domain in order to improve answer accuracy, relevance, and usability for expert-level question answering.

AM knowledge is distributed across heterogeneous sources such as academic literature, manufacturer documentation, technical standards, and procedural guides. Although general LLMs demonstrate strong linguistic capabilities, they frequently fail to retrieve and contextualize such domain-specific information. Two common approaches to address this limitation are domain-specific fine-tuning and retrieval-augmented generation (RAG).

We construct a curated AM corpus and evaluate three configurations based on LLaMA-3-8B: (1) the pretrained baseline model, (2) a RAG system that retrieves relevant document chunks from a vector database, and (3) a model fine-tuned on raw domain text. Performance is evaluated using 200 expert-designed AM questions assessed by mechanical engineering experts for accuracy, relevance, and overall preference.

Results show that the RAG model consistently outperforms the baseline. Among the 200 questions, 75.5% of RAG responses are judged more accurate, 85.2% are preferred overall, and 90.8% are rated more relevant than baseline responses. In contrast, fine-tuning on raw AM text reduces performance, producing more accurate answers in only 5.6% of cases and more relevant answers in 32.5% of cases.

These results indicate that retrieval-augmented approaches provide a more effective pathway for adapting LLMs to specialized engineering domains than naive fine-tuning on unstructured technical data.

## Introduction

Large language models (LLMs) such as OpenAI's GPT-3 and GPT-4, Google's PaLM, and Meta's LLaMA have demonstrated remarkable capabilities in natural language understanding and generation [1]. These models, trained on massive internet-scale datasets, perform well on general-purpose benchmarks such as text generation, summarization and open-domain question answering. Scaling these models further has been shown to significantly improve their performance across such tasks. GPT-3, for example, achieved strong results in translation, open-domain question answering and even basic arithmetic-all without task-specific tuning [2]. However, despite their broad knowledge, general-purpose LLMs often struggle in highly specialized domains. When applied to fields such as additive manufacturing (AM), medical science or scientific research, these standard LLMs frequently fail to deliver reliable and accurate results.

Several studies highlight this performance gap. For example, GPT-3 was evaluated across 57 subjects in the MMLU benchmark and was found to underperform on technical topics, sometimes barely exceeding random guessing on niche domains [3]. In the field of AM, Chandrasekhar et al. [4] observed that GPT-4 provided only high-level responses, lacking detailed process understanding, while a retrieval-augmented domain-specific model (AMGPT) offered significantly more accurate answers. This was further confirmed by Halsey et al. [5], who found that GPT-4 could not provide manufacturing-specific details without augmentation, whereas a retrieval-augmented generation (RAG) based model significantly improved answer specificity and factuality. Eslaminia et al. [6] also emphasized that fused deposition modeling (FDM) tasks require specialized interdisciplinary knowledge that general-purpose LLMs fail to capture; they introduced the FDM-Bench to benchmark LLMs on realistic AM scenarios and found limited generalizability without domain adaptation. Similarly, Pak and Farimani [7] demonstrated that fine-tuning general LLMs on AM defect data (AdditiveLLM) led to substantial performance gains, suggesting that pretrained models alone cannot handle defect classification tasks reliably in AM settings. In clinical applications, DocOA-a specialized assistant built on the GPT-4 architecture demonstrated superior performance in osteoarthritis management compared to general-purpose LLMs, which often failed to incorporate context-specific medical guidelines [8]. In biomedical NLP, Chen et al. [9] showed that zero-shot GPT and LLaMA models consistently underperformed compared to domain-trained models like BioBERT across most benchmarks, particularly in information extraction. These findings collectively show that without domain adaptation, general LLMs often fail to meet expert-level performance standards in specialized tasks.

To bridge this gap, researchers have explored two primary strategies for domain adaptation: fine-tuning and RAG [10-15]. Fine-tuning involves updating a model's parameters using domain-specific information to embed relevant knowledge into the model itself. While this can enhance performance, it is computationally expensive and prone to overfitting, especially when the training data is noisy or limited [16]. Methods like LoRA and other parameter efficient fine-tuning (PEFT) techniques help to reduce adaptation costs yet still depend on the model internalizing domain knowledge to address gaps observed in zero-shot scenarios [17]. Furthermore, fine-tuned models are static, they cannot easily incorporate new information without retraining and often lack transparency regarding the sources of their claims.

RAG offers a more flexible and scalable alternative. Instead of storing all domain knowledge in model weights, RAG enables the LLM to retrieve relevant documents from a curated knowledge base during inference [18]. This retrieved content is then provided as input context, grounding the model's responses in actual reference material. Studies have shown that this approach improves factual accuracy, reduces hallucinations and allows models to stay current without needing retraining [19, 20]. For instance, DeepMind's RETRO model achieved performance on par with GPT-3 using 25 times fewer parameters by leveraging a large-scale retrieval mechanism [19]. In domains where correctness, traceability, and up-to-date knowledge are critical such as engineering and medical science, RAG provides a solid foundation for developing expert-level AI systems.

In this study, we investigate how different adaptation strategies influence the performance of a general-purpose LLM in AM question answering. To support this analysis, we first construct a curated AM dataset tailored for both model development and evaluation. Using this dataset, we compare three approaches- a pretrained baseline, a fine-tuned model and a RAG system to understand how each method handles domain-specific engineering queries. Through this comparative study, we examine the effectiveness of retrieval-based knowledge integration and the impact of fine-tuning on raw technical data. Our findings show that retrieval-based approaches provide a more reliable and scalable way to improve model performance in specialized domains, while also revealing important limitations of directly fine-tuning on unstructured AM data. These insights offer practical guidance for developing more accurate and trustworthy AI systems in AM.

## 2. Methodology

This section outlines the complete methodological framework used to adapt and evaluate a general-purpose LLM in the AM domain. The methodology integrates domain-specific dataset construction, parameter-efficient model adaptation, retrieval-based knowledge augmentation and expert-driven evaluation.

The overall workflow includes four major stages: domain knowledge collection, benchmark dataset construction, model adaptation strategies and expert-based evaluation. The high-level architecture of the proposed framework is illustrated in Figure 1.

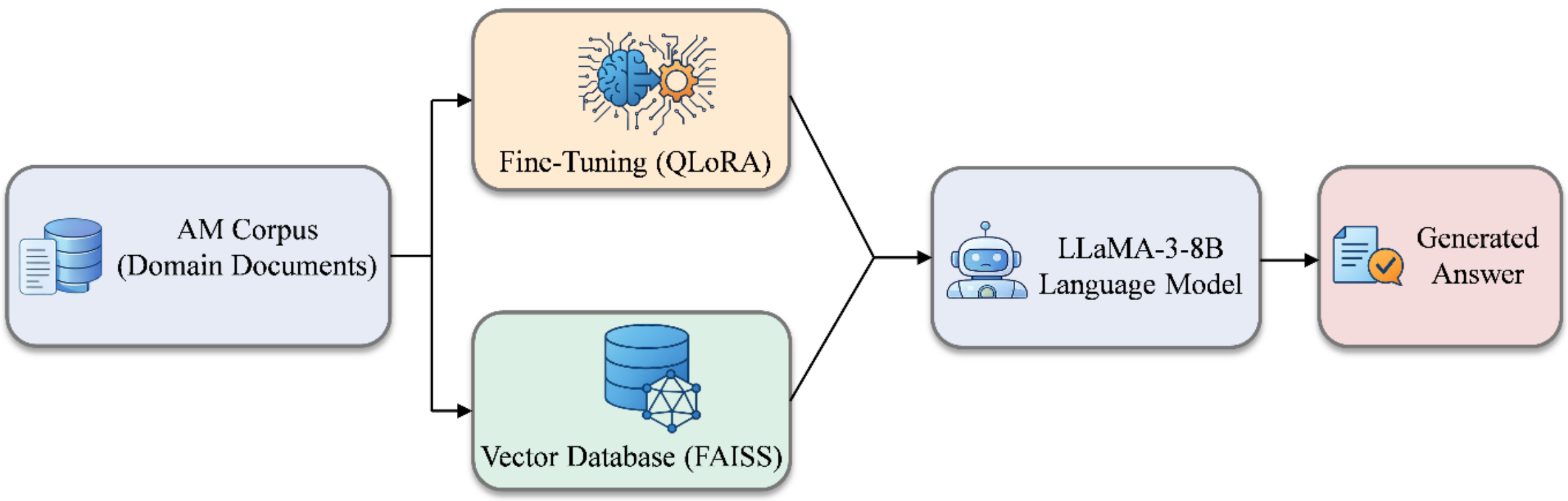

Figure 1: Overall methodology architecture of the proposed framework

As shown in Figure 1, domain-specific AM documents are first collected to build the AM corpus. The corpus is then used for both fine-tuning the LLM through QLoRA adaptation and constructing a vector database for RAG. The outputs from these components interact with the LLaMA-3-8B language model to produce the final generated responses.

**2.1 Data Collection**

To provide the model with relevant domain knowledge, a specialized AM corpus was constructed from multiple sources. These include approximately 250+ peer-reviewed research articles, 30+ book chapters, 150+ AM process guidelines, 50+ international technical standards (e.g., ASTM, ISO) and SOPs from industrial sources and 120+ material datasheets. These materials covered different AM techniques such as FDM, stereolithography (SLA), digital light processing (DLP), directed energy deposition (DED) etc., also subdomains such as material science, machine settings, pre-processing, post-processing, design rules, failure analysis etc. These techniques and subdomains were identified based on initial literature review and keyword based search which highlighted the most frequently studied and practically relevant areas in AM domain.

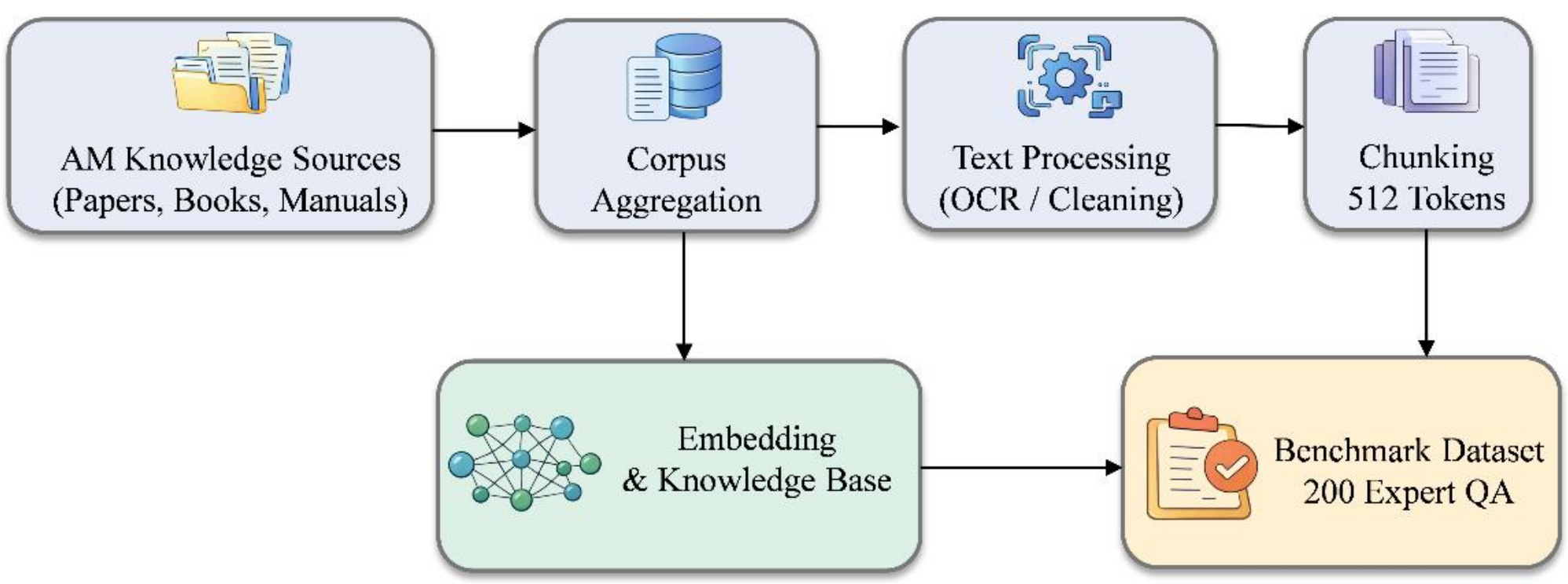

Figure 2: Data collection and preprocessing pipeline for AM corpus construction.

As illustrated in Figure 2, the data collection process begins by aggregating AM knowledge sources such as academic papers, books and technical manuals into a unified corpus. The data acquisition was carried out through a combination of systematic literature search and manual curation using academic databases such as Google Scholar, Scopus, and publisher repositories. Academic publications were collected using keyword-based searches (e.g., “polymer composite,” “digital light processing (DLP),” “direct ink writing (DIW),” "Ceramic AM" etc.) across digital libraries. Technical manuals, standards and industrial documents were obtained from publicly available repositories and manufacturer websites. The collection and curation process was performed by a team of four researchers (led by Dr. Joyee), who manually reviewed and filtered the documents to ensure relevance to AM processes, polymer composite materials and design practices. Duplicate and low-quality documents were removed during this stage. The selected documents were then processed through a text preprocessing pipeline that includes optical character recognition (OCR), LaTeX text extraction and text cleaning to obtain machine-readable text. OCR was performed using standard PDF to text extraction tools, while LaTeX-based documents were parsed using automated scripts. Text cleaning included removal of formatting artifacts, symbols and irrelevant metadata.

The processed text is subsequently segmented into smaller passages using a chunking strategy with fixed token lengths in order to support efficient document retrieval. Each text chunk is converted into a dense vector representation using a sentence-embedding model based on the Sentence-Transformers framework, which has been widely used for semantic similarity and information retrieval tasks in natural language processing (NLP) [21].

The resulting embeddings are stored in a vector database constructed using Facebook AI Similarity Search (FAISS), a similarity search library designed for efficient nearest-neighbor retrieval in large-scale embedding spaces [22]. In parallel with corpus construction, a benchmark dataset consisting of 200 expert-designed question-answer pairs are manually developed. These questions cover important AM knowledge areas including materials selection, process parameter optimization, defect diagnosis and design guidelines. The questions vary in complexity in order to simulate realistic engineering queries that may arise during AM process planning or troubleshooting.

### 2.2 Model Architecture and Foundation

The base model used in this study is LLaMA-3-8B, a transformer-based large language model that provides strong natural language understanding and reasoning capabilities [23]. Instead of training a model from scratch, the pretrained model is adapted to the AM domain using parameter-efficient adaptation techniques.

To maintain computational efficiency, the majority of the pretrained model parameters remain frozen while a small subset of trainable parameters is introduced through parameter-efficient adaptation layers. This strategy enables the integration of specialized manufacturing knowledge without requiring large-scale retraining of the entire network.

In addition to the base transformer architecture, the framework integrates domain-specific knowledge through structured AM documents and retrieval-based reasoning mechanisms. These mechanisms allow the model to better capture relationships between materials, manufacturing processes and numerical process parameters. As a result, the model can reason about engineering constraints such as valid parameter ranges, material compatibility and manufacturing process limitations.

### 2.3 Adaptation Strategies

To analyze the effectiveness of different approaches for incorporating domain knowledge, three model configurations were implemented and compared. The experimental comparison setup is illustrated in Figure 3.

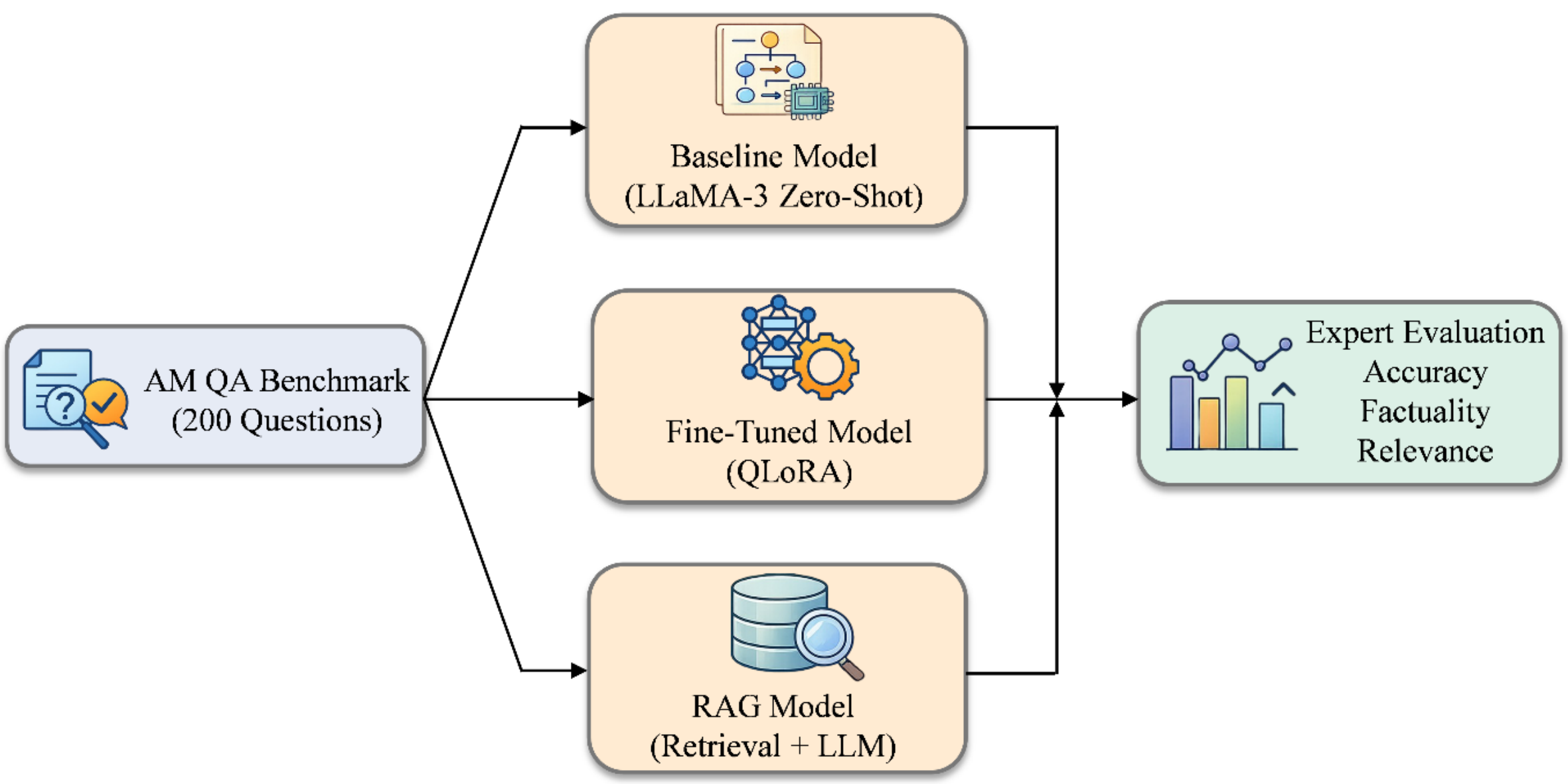

Figure 3: Experimental setup for model comparison. The benchmark dataset is evaluated across three configurations: baseline, fine-tuned, and RAG models; followed by expert-based assessment of generated responses.

As illustrated in Figure 3, the AM benchmark dataset is evaluated across three different model configurations: baseline, fine-tuned and RAG models. The outputs generated by each model are then evaluated by domain experts according to predefined technical criteria.

### 2.3.1 Baseline (Zero-Shot Inference)

In this setup, the pretrained LLaMA-3-8B baseline configuration evaluates the pretrained foundation model without any domain-specific adaptation. In this setup, the model answers AM related questions using only the general knowledge learned during pretraining. This configuration establishes a reference point for assessing how effectively a general-purpose LLM can handle specialized AM-related questions without explicit domain training.

### 2.3.2 Fine-Tuned Model

In the fine-tuning configuration, the base model is adapted using the domain-specific dataset described in Section 2.1. The training process follows a supervised learning approach in which the model learns to generate appropriate responses to AM-related questions. To reduce computational cost and memory requirements, quantized low-rank adaptation (QLoRA) is used (4-bit quantization with LoRA adapters) for parameter-efficient fine-tuning [24]. QLoRA introduces small trainable low-rank matrices into the transformer attention layers while keeping the majority of the original model parameters unchanged. This approach significantly reduces the number of trainable parameters while maintaining strong performance [25, 26]. The parameters included in Table 1 are used for fine tuning.

| Key Training Parameters | |
|---|---|
| Epochs | 9 |
| Batch Size | 8 |
| Learning Rate | $1 \times 10^{-6}$ |
| Optimizer | AdamW |
| Early Stopping | Based on validation loss |

Table 1: Key training parameters used for QLoRA-based fine-tuning of the LLaMA-3-8B model.

As shown in Table 1, the model is trained for several epochs using the AdamW optimizer, and early stopping is applied based on validation loss to prevent overfitting due to the relatively small dataset size. This approach allows the model to embed AM knowledge directly into its internal representations, improving its ability to produce technically accurate responses

### 2.3.3 Retrieval-Augmented Generation (RAG)

The third configuration uses a RAG framework to enhance the language model with external domain knowledge. RAG combines neural language models with document retrieval systems, allowing the model to access relevant documents during inference instead of relying solely on internal model parameters [18, 27]. The RAG workflow consists of two stages: offline document indexing and online inference, which are illustrated in Figure 4.

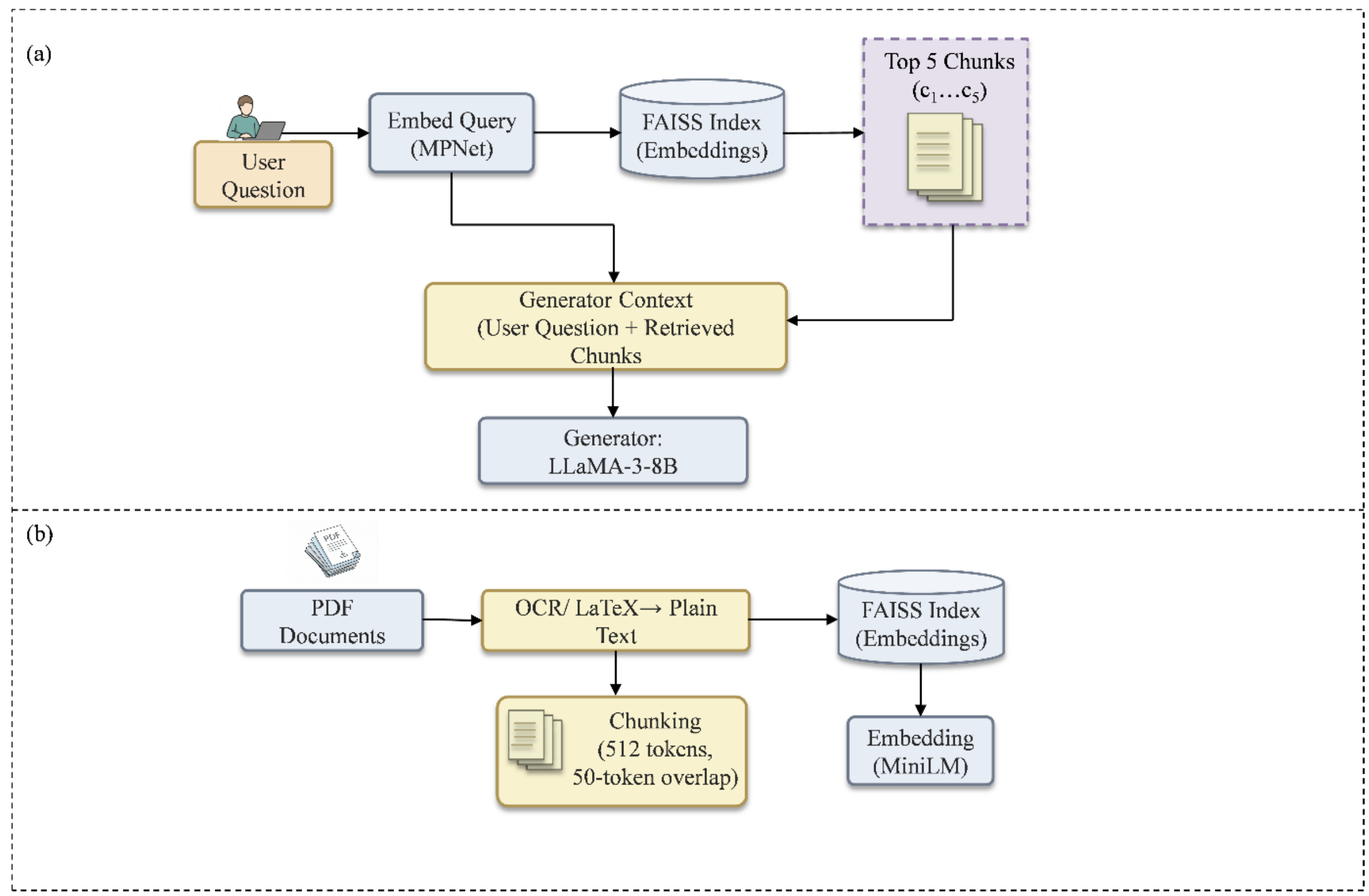

Figure 4: RAG workflow. (a) Online inference pipeline showing query embedding, document retrieval and response generation. (b) Offline indexing process including document preprocessing, chunking, and embedding storage in the vector database.

During the offline stage, domain documents are converted into machine-readable text through OCR or LaTeX extraction, as shown in Figure 4(b). The cleaned text is then divided into overlapping segments using a chunking strategy that preserves contextual continuity. Each text segment is converted into a semantic embedding using a sentence embedding model and stored in a FAISS vector database to enable efficient similarity-based retrieval [22].

Also, during the online inference stage, a user query is first converted into an embedding representation. The system retrieves the most relevant document chunks from the FAISS index using vector similarity search. These retrieved passages are then concatenated with the user query and provided as contextual input to the LLaMA-based language model. By grounding the generated responses in retrieved technical documents, the RAG approach improves factual accuracy and significantly reduces hallucination in technical question answering tasks.

## 2.4 Training Strategy and Knowledge Integration

The model training process follows a staged domain adaptation strategy. Initially, the model is exposed to general technical knowledge related to engineering and materials science to strengthen its understanding of technical vocabulary. Subsequently, training focuses on AM literature to capture relationships between process parameters, materials and performance outcomes.

Special attention is given to representing numerical parameters and process constraints, which are critical for manufacturing applications. The training procedure emphasizes learning valid parameter relationships such as temperature ranges, layer thickness values, curing conditions and process limits associated with different materials and AM processes. This improves the model's ability to provide realistic and technically consistent parameter recommendations.

### 2.5 Evaluation Protocol

The performance of the three model configurations were evaluated using a benchmark dataset consisting of 200 expert-designed AM questions. These questions were developed through a combination of literature review, research experience and practical teaching interactions. Specifically, the dataset was created by a team of four researchers, who curated questions based on commonly discussed topics in AM literature, frequently encountered research problems and real-world queries observed during classroom teaching and online technical forums.

The questions were designed to cover a broad range of AM knowledge areas, including materials selection, process parameter optimization, defect diagnosis, and design guidelines. In addition, the questions span multiple levels of complexity, ranging from fundamental concepts (e.g., "What is the typical layer height in FDM printing?") to more advanced process-level inquiries (e.g., "How does nozzle geometry influence extrusion behavior in DIW printing?"), ensuring that the benchmark reflects realistic engineering scenarios.

Each generated response is assessed using three evaluation criteria:

Accuracy: It measures whether the response is technically correct according to established AM knowledge, consistent with evaluation practices in question-answering benchmarks [28, 29].

Factuality: It evaluates whether the response avoids hallucinations and remains consistent with known engineering principles, which is a critical concern in modern LLM evaluation [30, 31].

Relevance: It assesses whether the response directly addresses the user's query and provides meaningful information, as commonly used in information retrieval and QA systems [32].

A panel of four AM experts evaluates the generated responses using a structured qualitative assessment framework. The evaluation is conducted using binary judgments (e.g., Yes/No) across the defined criteria, rather than numerical scoring. In addition, experts perform pairwise comparisons between responses from different model configurations to determine which answer better addresses the given query. This approach allows for a more intuitive and application-oriented assessment of model performance, reflecting how engineers evaluate the correctness and usefulness of technical information in real-world scenarios. Similar human judgment-based evaluation strategies have been widely adopted in recent LLM studies, particularly for domain-specific tasks where interpretability and practical reliability are critical [30].

### 2.6 Experimental Environment

All experiments are implemented using modern deep learning frameworks. The system uses the Hugging Face Transformers library for model implementation and parameter-efficient fine-tuning tools for QLoRA-based training. The retrieval system used in the RAG framework is implemented using FAISS for efficient vector similarity search. Training and inference experiments are conducted on GPU-accelerated computing hardware to ensure efficient model optimization and reproducibility of results.

## 3. Results

The entire evaluation focuses on comparative performance across accuracy, relevance and overall response preference.

### 3.1. Comparative Performance: RAG vs Baseline

The comparative performance of the RAG model and the pretrained baseline model is presented in Figure 5. The evaluation is based on three criteria: overall answer preference ("Better Answer"), accuracy and relevance, as assessed by domain experts.

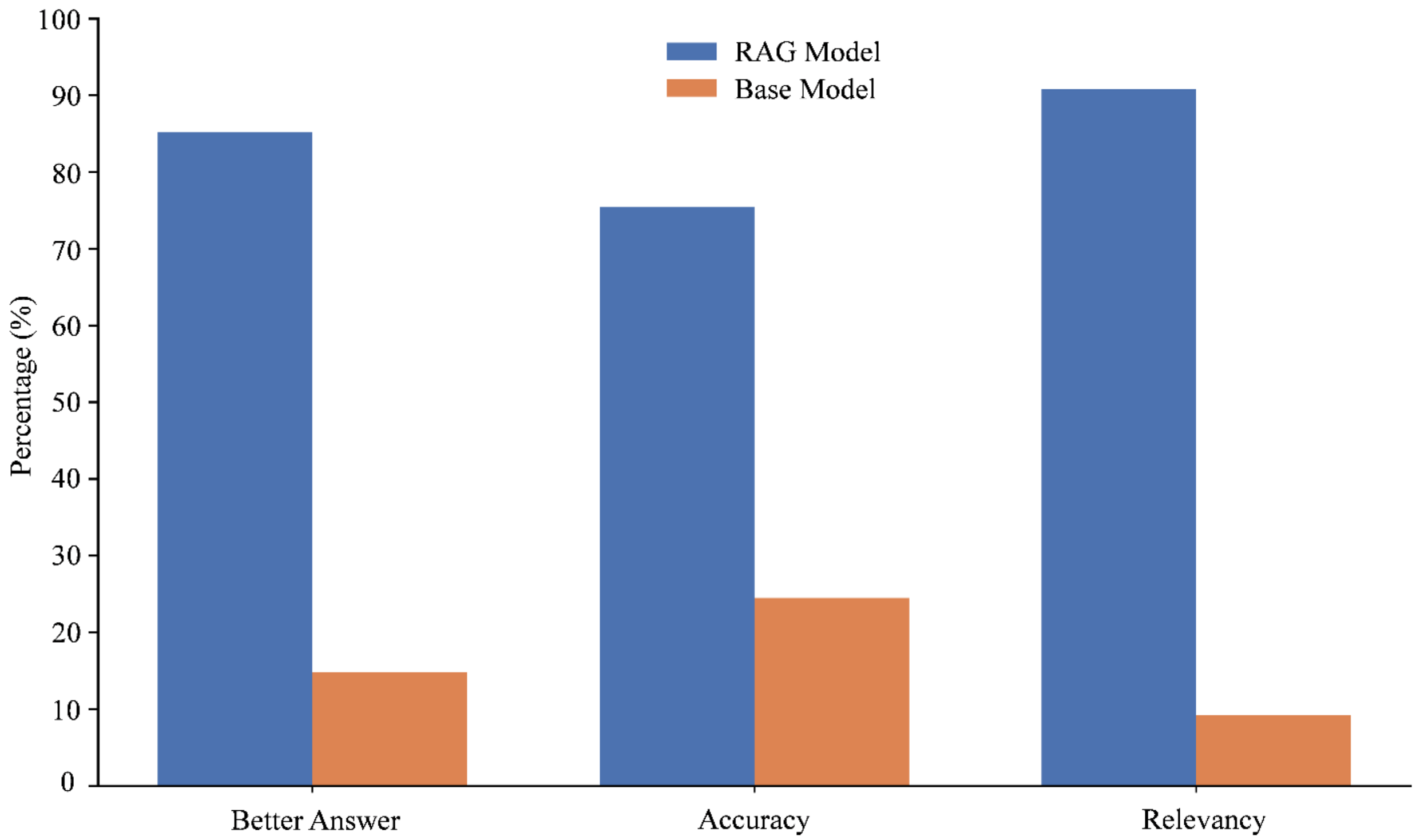

Figure 5: Comparative performance of RAG and baseline models across evaluation metrics.

As shown in Figure 5, the RAG model consistently outperforms the baseline model across all evaluation metrics. In terms of overall preference, 85.2% of the responses generated by the RAG model are judged to be better than those produced by the baseline model, whereas only 14.8% of baseline responses are preferred. This indicates a strong preference among evaluators for answers that are supported by retrieved domain-specific context.

A similar trend is observed in accuracy. The RAG model achieves an accuracy rate of 75.5%, significantly higher than the 24.5% achieved by the baseline model. This improvement suggests that incorporating retrieved knowledge helps the model generate technically correct responses by grounding its outputs in relevant AM information.

The difference is even more pronounced in relevance. The RAG model achieves a relevance score of 90.8%, compared to only 9.2% for the baseline model. This highlights the effectiveness of retrieval in ensuring that responses directly address the user's query and remain contextually aligned with the problem. Overall, the results show a clear improvement in performance when retrieval is incorporated into the model. The RAG configuration achieves higher scores across preference, accuracy and relevance compared to the pretrained baseline, indicating more consistent and contextually aligned responses in domain-specific question answering.

### 3.2 Performance of Fine-Tuned Model Compared to Baseline

The performance of the fine-tuned model is compared with the pretrained baseline model in Figure 6, focusing on two key evaluation metrics: accuracy and relevance. This comparison aims to assess whether domain-specific fine-tuning improves the model's ability to generate technically correct and contextually relevant responses.

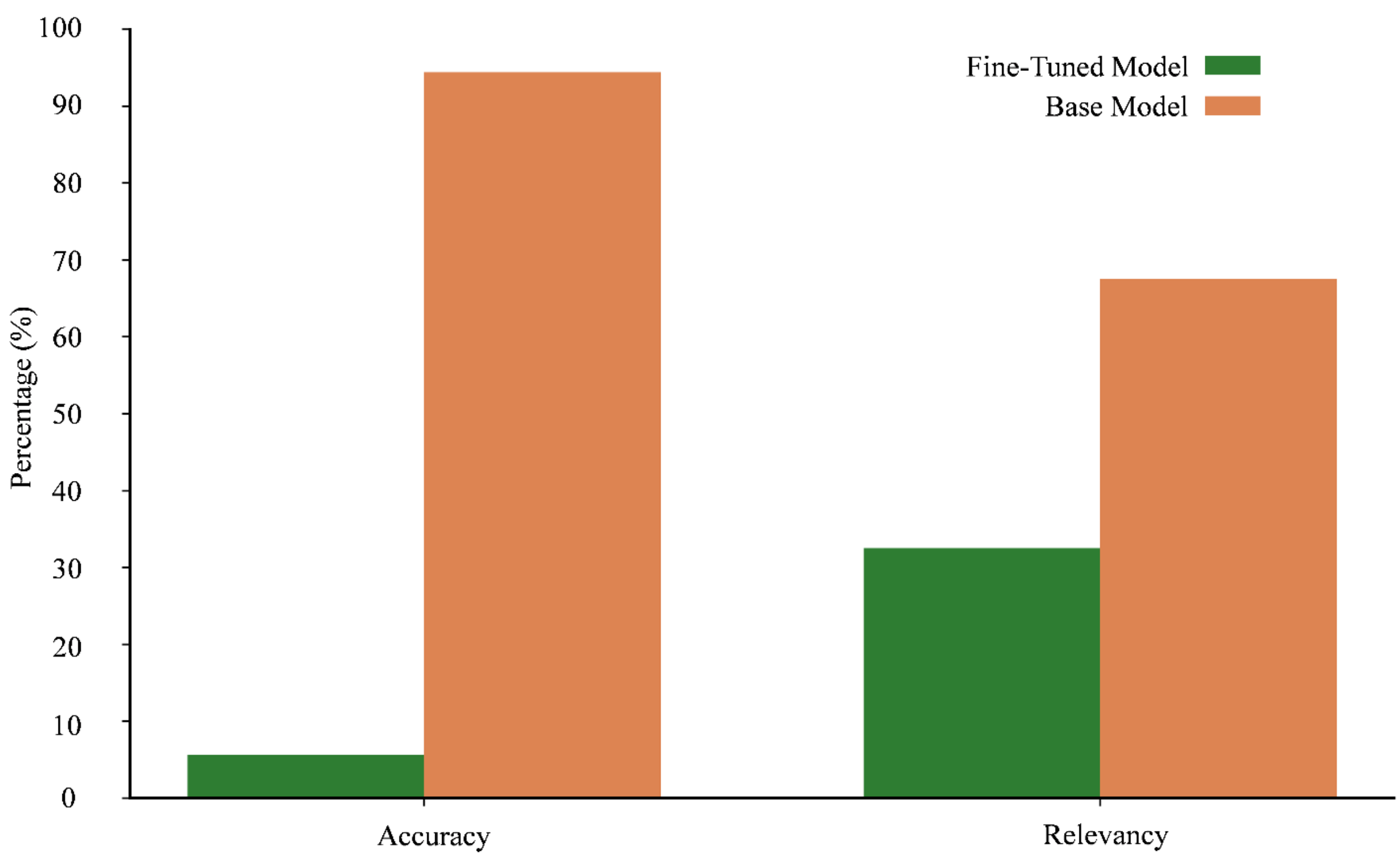

Figure 6: Performance comparison between fine-tuned and baseline models.

As shown in Figure 6, the fine-tuned model does not outperform the baseline model. In terms of accuracy, the fine-tuned model achieves only 5.6%, whereas the baseline model attains a significantly higher accuracy of 94.4%. This indicates that the fine-tuning process, when applied to raw and unstructured domain data, does not effectively enhance the model's ability to generate correct technical responses.

A similar trend is observed for relevance. The fine-tuned model achieves a relevance score of 32.5%, while the baseline model reaches 67.5%. This suggests that the fine-tuned model struggles to consistently produce responses that directly address the user's query, despite being exposed to domain-specific data during training. This result indicates that the fine-tuned model does not outperform the pretrained baseline in this task. The observed reduction in both accuracy and relevance suggests that the applied fine-tuning strategy does not effectively improve question-answering performance on the given dataset.

### 3.3 Information Quality and Usability Analysis

The quality and usability of responses generated by the RAG and fine-tuned models are further analyzed in Figure 7, which presents a comparative evaluation across multiple qualitative metrics, including harmfulness, contextual correctness, understandability and real-world applicability.

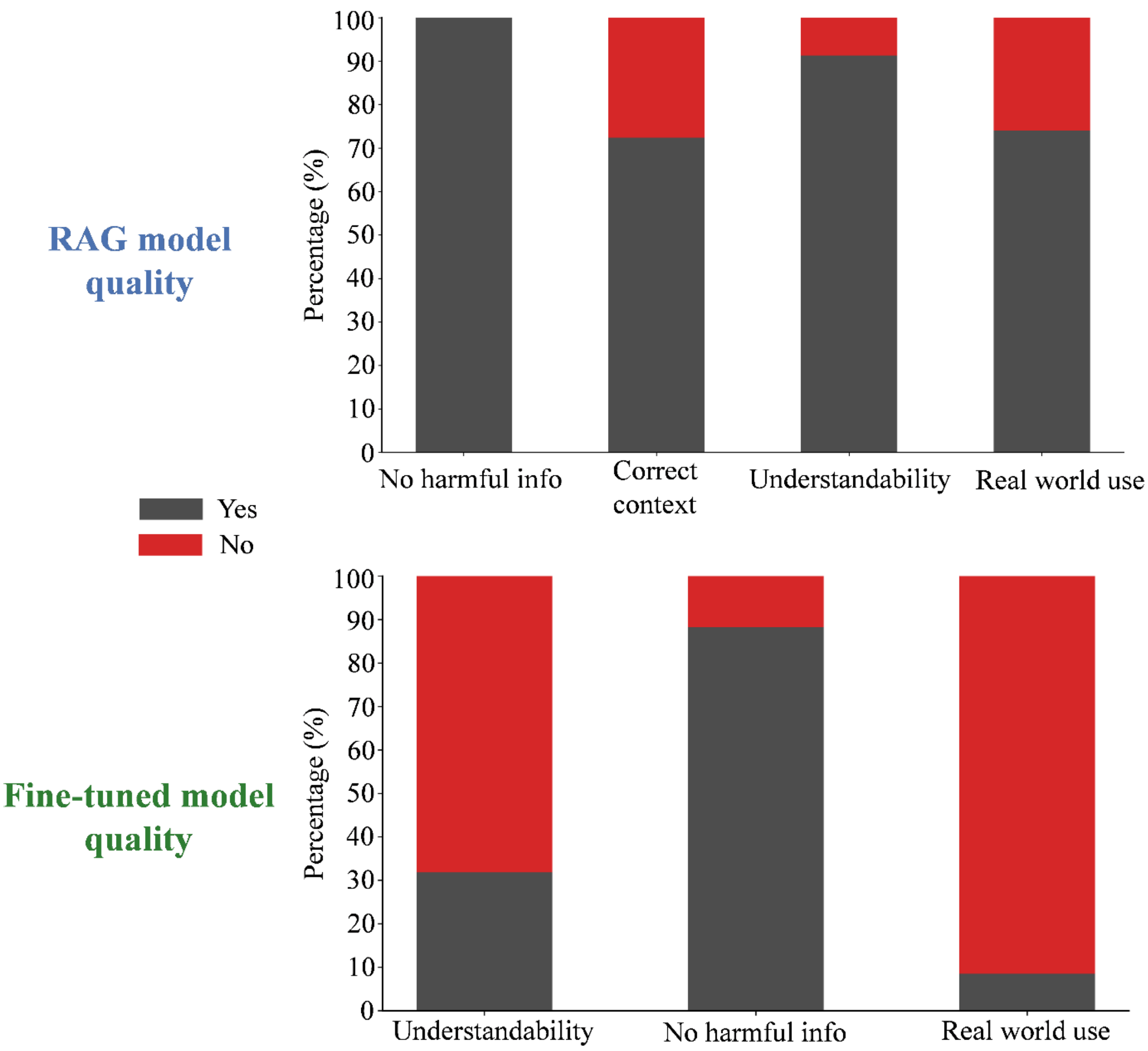

Fig 7: Information quality and usability analysis of RAG and fine-tuned models.

As illustrated in Figure 7, the RAG model demonstrates consistently strong performance across all evaluation dimensions. The model achieves 100% in producing non-harmful responses, indicating that all generated outputs are safe and aligned with acceptable technical standards. In terms of contextual correctness, 72.4% of the responses are evaluated as containing accurate and relevant information, while 27.6% exhibit some degree of inconsistency. The RAG model also performs well in understandability, with 91.3% of responses being clearly interpretable.

Additionally, 74.0% of the responses are considered applicable in real-world engineering contexts, suggesting that the model is capable generating practically useful guidance for domain specific tasks.

In contrast, the fine-tuned model exhibits significantly lower performance across the evaluated metrics, as shown in the lower panel of Figure 7. Only 31.8% of its responses are judged to be understandable, while the majority (68.2%) fail to clearly convey meaningful information. Although the model maintains a relatively high level of safety, with 88.2% of responses classified as non-harmful, its practical utility remains limited as only 8.5% of the responses are considered suitable for real-world use, indicating that most outputs lack actionable or technically reliable content. This result demonstrates a comparison highlights and distinction between retrieval-based and fine-tuning-based adaptation strategies. The RAG model consistently achieves higher scores across all evaluated metrics, including safety, contextual correctness, understandability and real-world applicability, whereas the fine-tuned model shows comparatively lower performance in these aspects.

## 4. Discussions

The results of this study demonstrate that retrieval-based augmentation is significantly more effective than direct fine-tuning for adapting large language models to specialized engineering domains. The RAG-enhanced model consistently outperforms both the baseline and fine-tuned configurations across all evaluation criteria, including accuracy, relevance and overall response quality.

These findings are consistent with prior work in retrieval-augmented generation, where integrating external knowledge sources has been shown to improve factual accuracy and reduce hallucinations in knowledge-intensive tasks [19]. In particular, the strong improvement in relevance observed in this study aligns with the core principle of RAG systems, which retrieve contextually similar documents to guide response generation.

In contrast, the observed performance degradation in the fine-tuned model highlights a known limitation of parameter-based adaptation when applied to unstructured data. Similar observations have been reported in prior studies, where fine-tuning on raw domain text without task-specific supervision can lead to overfitting or loss of general reasoning capabilities [33, 34]. This suggests that simply exposing a model to domain knowledge is insufficient for improving downstream task performance, especially in question-answering settings.

Another key advantage of the RAG approach is its flexibility and scalability. Unlike fine-tuned models, which require retraining to incorporate new knowledge, RAG systems can update their knowledge base dynamically by adding new documents to the retrieval index. This property is particularly important in engineering domains such as additive manufacturing, where knowledge evolves rapidly and includes both theoretical and procedural components. Furthermore, the strong performance of RAG in real-world applicability and contextual correctness suggests that retrieval-based grounding plays a critical role in bridging the gap between general-purpose language models and domain-specific expertise. This aligns with recent domain-specific LLM studies in manufacturing and biomedical fields, where retrieval mechanisms have been shown to significantly improve reliability and trustworthiness of generated outputs. Overall, the findings of this study reinforce the growing consensus that hybrid approaches combining language models with external knowledge retrieval provide a more effective pathway for deploying LLMs in specialized technical domains.

Despite the promising results, several limitations should be acknowledged. First, the fine-tuning process was conducted using raw textual data without incorporating structured question-answer pairs. This limits the model's ability to learn how to generate high-quality responses in a QA setting. Future improvements may require curated QA datasets for more effective supervised fine-tuning. Second, the evaluation relies on expert judgment, which may introduce subjectivity. Although multiple experts were involved, variations in interpretation and experience could influence evaluation outcomes. Third, the size of the benchmark dataset is relatively limited (200 questions), which may not fully capture the diversity of real-world AM queries. Expanding the dataset could provide more comprehensive evaluation. Additionally, the RAG system depends heavily on the quality and coverage of the underlying corpus. Missing or incomplete information in the knowledge base may lead to incomplete or suboptimal responses. Finally, the study focuses primarily on textual knowledge representation. Numerical reasoning and simulation-based validation, which are important in engineering applications, are not explicitly addressed in this work.

## 5. Conclusion and Future Work

This study investigates the effectiveness of different adaptation strategies for applying a general-purpose LLM to additive manufacturing question answering. The results show that retrieval-augmented generation significantly improves model performance by grounding responses in domain-specific knowledge. In contrast, fine-tuning on raw textual data does not consistently enhance performance and may degrade the model's ability to generate accurate and relevant answers. These findings highlight the importance of incorporating external knowledge sources when deploying LLMs in specialized engineering domains. Retrieval-based approaches provide a flexible and scalable solution that can adapt to evolving knowledge without requiring costly retraining.

For future work, several directions can be explored. First, developing a larger and more structured domain-specific QA dataset could improve fine-tuning effectiveness. Second, hybrid approaches that combine retrieval and supervised fine-tuning may further enhance performance. Third, integrating numerical reasoning capabilities and simulation-based validation could improve the applicability of LLMs in engineering decision-making. Overall, this work demonstrates that retrieval-based methods offer a practical pathway for building reliable AI assistants in AM and similar technical domains.